\documentclass[10pt,twocolumn,letterpaper]{article}

\usepackage{cvpr}              

\usepackage{times}
\usepackage{epsfig}
\usepackage{graphicx}
\usepackage{amsmath}
\usepackage{amssymb}
\usepackage{color, colortbl}
\usepackage{multirow}
\usepackage{booktabs}
\usepackage{float}

\usepackage[accsupp]{axessibility}

\usepackage[pagebackref=true,breaklinks=true,colorlinks,bookmarks=false]{hyperref}

\def\cvprPaperID{7619} 
\def\confYear{CVPR 2022}
\definecolor{Gray}{gray}{0.9}

\begin{document}

\title{Exploring Structure-aware Transformer over Interaction Proposals\\ for Human-Object Interaction Detection\thanks{{\small This work was performed at JD Explore Academy.}}}

\author{Yong Zhang$^{\dag}$, Yingwei Pan$^{\ddag}$, Ting Yao$^{\ddag}$, Rui Huang$^{\dag}$, Tao Mei$^{\ddag}$, and Chang-Wen Chen$^{\S}$\\
{\normalsize\centering$^{\dag}$ The Chinese University of Hong Kong, Shenzhen}~
{\normalsize\centering$^{\ddag}$ JD Explore Academy}\\
{\normalsize\centering$^{\S}$ The Hong Kong Polytechnic University}\\
{\tt\small yongzhang@link.cuhk.edu.cn, \{panyw.ustc, tingyao.ustc\}@gmail.com, ruihuang@cuhk.edu.cn,}\\
{\tt\small tmei@jd.com, changwen.chen@polyu.edu.hk}
}

\maketitle

\begin{abstract}
Recent high-performing Human-Object Interaction (HOI) detection techniques have been highly influenced by Transformer-based object detector (i.e., DETR). Nevertheless, most of them directly map parametric interaction queries into a set of HOI predictions through vanilla Transformer in a one-stage manner. This leaves rich inter- or intra-interaction structure under-exploited. In this work, we design a novel Transformer-style HOI detector, i.e., Structure-aware Transformer over Interaction Proposals (STIP), for HOI detection. Such design decomposes the process of HOI set prediction into two subsequent phases, i.e., an interaction proposal generation is first performed, and then followed by transforming the non-parametric interaction proposals into HOI predictions via a structure-aware Transformer. The structure-aware Transformer upgrades vanilla Transformer by encoding additionally the holistically semantic structure among interaction proposals as well as the locally spatial structure of human/object within each interaction proposal, so as to strengthen HOI predictions. Extensive experiments conducted on V-COCO and HICO-DET benchmarks have demonstrated the effectiveness of STIP, and superior results are reported when comparing with the state-of-the-art HOI detectors. Source code is available at \url{https://github.com/zyong812/STIP}.
\end{abstract}

\begin{figure}
\vspace{-0.4in}
\begin{center}
   \includegraphics[width=0.83\linewidth]{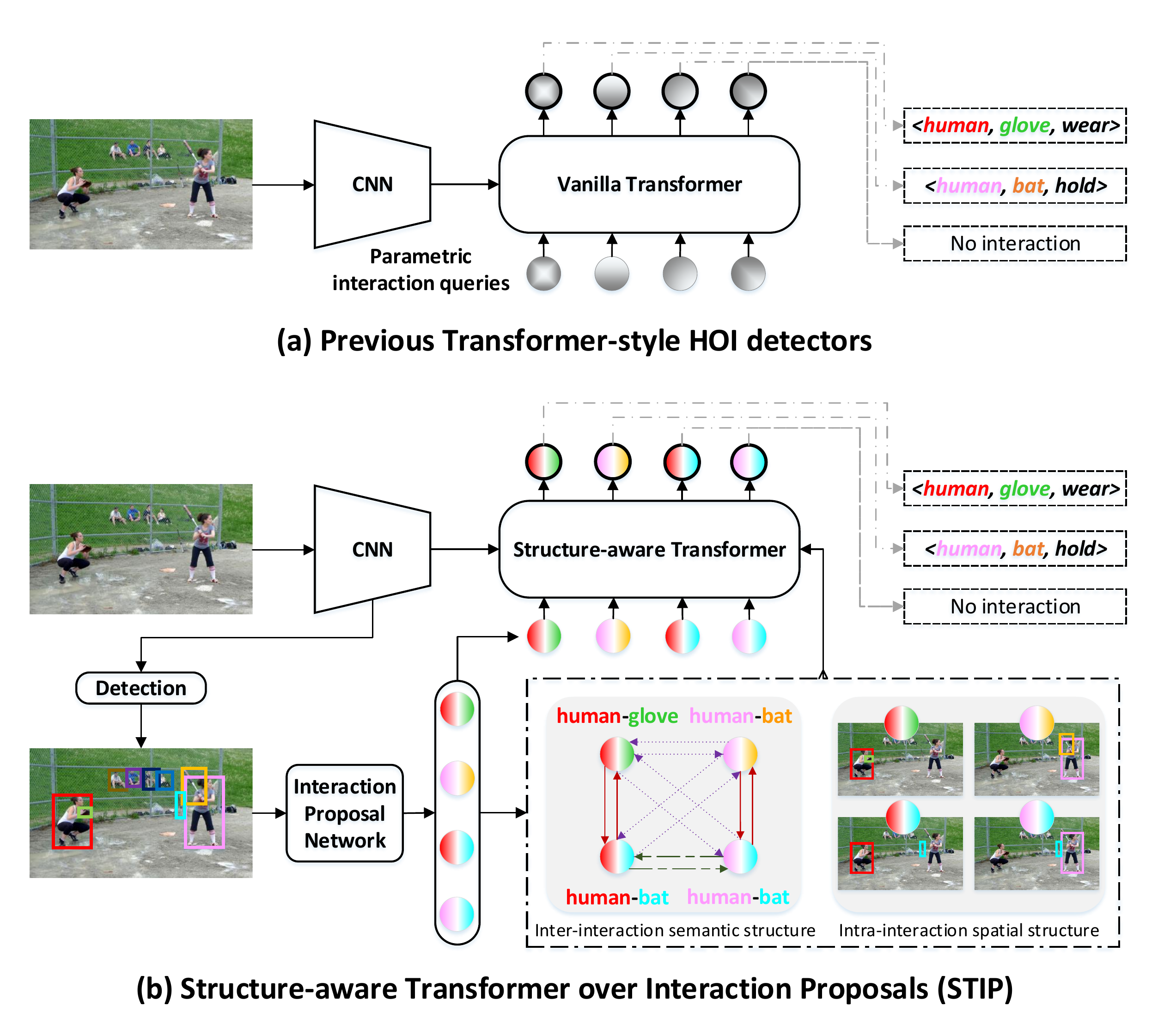}
\end{center}
\vspace{-0.3in}
   \caption{Comparison between existing Transformer-style HOI detectors and our STIP. (a) Existing Transformer-style HOI detectors directly transform the parametric interaction queries into HOI predictions via vanilla Transformer in a one-stage fashion. (b) STIP adopts a two-phase solution, i.e., first producing interaction proposals via Interaction Proposal Network, and then mapping the non-parametric interaction queries (i.e., interaction proposals) into HOI predictions. Both of the inter- and intra-interaction structures derived from interaction proposals are additionally exploited to boost HOI set prediction through a structure-aware Transformer.}
\label{fig:overview}
\vspace{-0.3in}
\end{figure}

\section{Introduction}

Human-Object Interaction (HOI) detection \cite{chao2015hico,gupta2015visual} is intended to localize the interactive human-object pairs within an image and identify the interactions in between, yielding the HOI predictions in the form of {\emph{$\langle human, object, interaction\rangle$}} triplets. Practical HOI detection systems perform the human-centric scene understanding, and thus have a great potential impact for numerous applications, such as surveillance event detection \cite{adam2008robust,dogariu2020human} and robot imitation learning \cite{argall2009survey}.
In general, conventional HOI detectors \cite{gao2020drg,gao2018ican,gkioxari2018detecting,kim2020uniondet,liu2020amplifying,liao2020ppdm,qi2018learning,wang2020contextual,wang2019deep,wang2020learning} tackle the HOI set prediction task in an \textbf{indirect} way, by formalizing it as surrogate regression and classification problems for human/object/interaction. Such indirect approach needs a subsequent post-processing by collapsing near-duplicate predictions and heuristic matching \cite{kim2020uniondet,liao2020ppdm,wang2020learning}, and thus cannot be trained in an end-to-end fashion, resulting in a sub-optimal solution.
The intent to overcome the problem of sub-optimal solution leads to the development of recent state-of-the-art HOI detectors \cite{chen2021reformulating,kim2021hotr,tamura2021qpic,zou2021end} that follow the Transformer-based detector of DETR \cite{carion2020end} to cast HOI detection as a \textbf{direct} set prediction problem, and embrace the ``end-to-end'' philosophy (Figure \ref{fig:overview} (a)). In particular, a vanilla Transformer is commonly utilized to map the parametric interaction queries (i.e., the learnable positional embedding sequence) into a set of HOI predictions in a one-stage manner.
However, these HOI detectors start the HOI set prediction from the parametric interaction queries with randomly initialized embeddings. That is, the correspondence between parametric interaction queries and output HOIs (commonly assigned by Hungarian algorithm for training) is \textbf{dynamic} in which the interaction query corresponding to each target HOI (e.g., ``human hold bat") is unknown at the beginning of HOI set prediction. This can adversely hinder the exploration of prior knowledge (i.e., \textbf{inter-interaction} or \textbf{intra-interaction structure}) which would be very useful for relational reasoning among interactions in HOI set prediction.

Specifically, by inter-interaction structure, we refer to the holistic semantic dependencies among HOIs, which can be~directly defined by considering whether or not two HOIs share the same human or object. Such structure implies the common sense knowledge that shall facilitate the prediction of one HOI by exploiting its semantic dependencies against other HOIs. Taking the input image in Figure \ref{fig:overview} as an example, the existence of ``human wear (baseball) glove" provides strong indication for ``(another) human hold bat".~Moreover, the intra-interaction structure can be interpreted~as the local spatial structure for each HOI, i.e., the layout of human and object, which acts as additional prior knowledge to steer model's attention over image areas for depicting the interaction.

In this work, we design a novel scheme based on Transformer-style HOI detector, namely Structure-aware Transformer over Interaction Proposals (STIP). The design innovation is to decompose the one-stage solution of set prediction into two cascaded phases, i.e., first producing the interaction proposals (i.e., the probably interactive human-object pairs) and then performing HOI set prediction based on the interaction proposals (Figure \ref{fig:overview} (b)). By taking the interaction proposals derived from Interaction Proposal Network (IPN) as non-parametric interaction queries, STIP naturally triggers the subsequent HOI set prediction with more reasonable interaction queries, leading to \textbf{static} query-HOI correspondence that capable of boosting HOI set prediction. As a beneficial by-product, the predicted interaction proposals offer a fertile ground for constructing a structured understanding across interaction proposals or between human \& object within each interaction proposal. A particular form of Transformer, i.e., structure-aware Transformer, is designed accordingly to encode the inter-interaction or intra-interaction structure for enhancing HOI predictions.

In sum, we have made the following contributions: (1) The proposed two-phase implementation of Transformer-style HOI detector is shown capable of seamless incorporation of potential interactions among HOI proposals to overcome the problem associated with one-stage approach; (2) The exquisitely designed structure-aware Transformer is shown able to facilitate additional exploitation opportunity for utilizing inter-interaction and intra-interaction structure for enhanced performance of the vanilla Transformer; (3) The proposed structure-aware Transformer approach has been properly analyzed and verified through extensive experiments over V-COCO and HICO-DET datasets to validate its potential in solving the problems associated with one-stage approach to achieve desirable HOI detection.

\section{Related Work}

The task of Human-Object Interaction (HOI) detection has been primordially defined \cite{chao2015hico,gupta2015visual} and recent developments of HOI detectors can be briefly divided into two categories: the two-stage methods and one-stage approaches.

\textbf{Two-stage Methods.}
The first category schemes \cite{chao2018learning,gao2020drg,gao2018ican,gkioxari2018detecting,hou2021affordance,hou2021detecting,li2020hoi,li2019transferable,liu2020amplifying,liu2020consnet,qi2018learning,ulutan2020vsgnet,wang2020contextual,wang2019deep,zhong2020polysemy,zhang2021mining} mainly adopt two-stage paradigm, i.e., first detect humans/objects via off-the-shelf modern object detectors (e.g., Faster R-CNN \cite{ren2015faster}) and then carry out interaction classification. A number of schemes have been proposed to strengthen the HOI feature learning in the second stage for interaction classification. Generally, similar to prior works for visual relationship detection \cite{li2017vip,mi2020hierarchical,yao2018exploring,zhang2022boosting,zhang2019graphical}, HOI features are typically derived from three perspectives \cite{chao2018learning,gao2018ican,gkioxari2018detecting}: appearance/visual features of humans and objects, spatial features (e.g., the pairwise bounding boxes of human-object pair), and linguistic feature (e.g., the semantic embeddings of human/object labels). Various approaches \cite{gao2020drg,he2021exploiting,qi2018learning,ulutan2020vsgnet,wang2020contextual,zhang2020spatially} further capitalize on message passing mechanism to perform relational reasoning over instance-centric graph structure, aiming to enrich HOI features with global contextual information among human and object instances. The authors in \cite{wang2019deep} devise contextual attention mechanism to facilitate the mining of contextual cues. Moreover, the information about human pose \cite{gupta2019no,li2019transferable,zhong2020polysemy}, body parts \cite{zhou2019relation} or detailed 3D body shape \cite{li2020detailed} can also be exploited to enhance HOI feature representation. In \cite{liu2020consnet,xu2019learning}, additional knowledge from external source and language domain are further exploited to boost HOI feature learning. Most recently, the ATL scheme \cite{hou2021affordance} constructs the affordance feature bank across multiple HOI datasets and injects affordance feature into object representations when inferring interactions.

\begin{figure*}[!tb]
\vspace{-0.36in}
\begin{center}
   \includegraphics[width=0.8\linewidth]{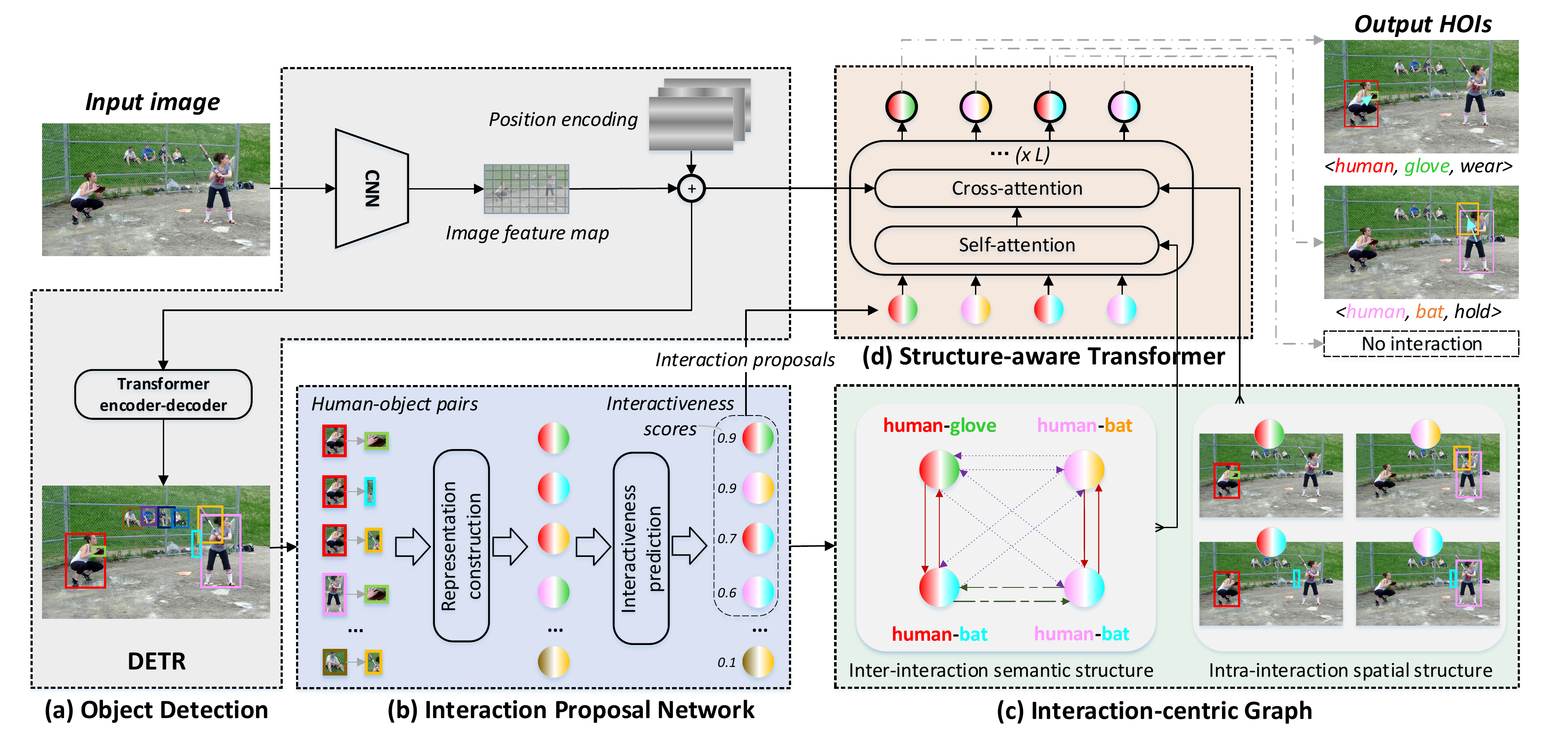}
   \vspace{-0.28in}
\end{center}
   \caption{An overview of our proposed STIP framework. (a) Given an input image, we adopt an off-the-shelf DETR to detect the human and object instances within this image. (b) Based on the detected human and object instances, the Interaction Proposal Network (IPN) constructs all possible human-object pairs and then predicts the interactiveness score of each human-object pair. The most interactive human-object pairs with highest interactiveness scores are taken as the output interaction proposals. (c) Next, by taking all interaction proposals as graph nodes and exploiting semantic connectivity as edges, we build an interaction-centric graph that unfolds rich inter-interaction semantic structure and intra-interaction spatial structure. (d) Finally, a structure-aware Transformer is utilized to transform the non-parametric interaction queries (i.e., interaction proposals) into a set of HOI predictions by additionally guiding relational reasoning with the inter- or intra-interaction structure derived from interaction-centric graph.}
\label{fig:framework}
\vspace{-0.2in}
\end{figure*}

\textbf{One-stage Approaches.}
The second category schemes mainly construct one-stage HOI detectors \cite{chen2021reformulating,kim2020uniondet,kim2021hotr,liao2020ppdm,tamura2021qpic,wang2020learning,zhong2021glance,zou2021end} by directly predicting HOI triplets, which are potentially faster and simpler than two-stage HOI detectors.
UnionDet \cite{kim2020uniondet} is one of first attempts that directly detects the union regions of human-object pairs in a one-stage manner. Other schemes \cite{liao2020ppdm,wang2020learning} formulate HOI detection as a keypoint detection problem, and thus enable one-stage solution for this task. Most recently, inspired by the success of Transformer-based object detectors (e.g., DETR \cite{carion2020end}), there has been a steady momentum of breakthroughs that push the limits of HOI detection by using Transformer-style architecture. In particular, the authors in \cite{tamura2021qpic,zou2021end} employ a single interaction Transformer decoder to predict a set of HOI triplets, and the whole architecture can be optimized in an end-to-end fashion with Hungarian loss. However, the authors in \cite{chen2021reformulating,kim2021hotr} design two parallel Transformer decoders for detecting interactions and instances, and the outputs are further associated to produce final HOI predictions.

\textbf{This Scheme.} The proposed STIP can also be considered as Transformer-style architecture that tackles HOI detection as a set prediction problem, which eliminates the post-processing and enables the architecture to be end-to-end trainable. Unlike existing Transformer-style methods \cite{chen2021reformulating,kim2021hotr,tamura2021qpic,zou2021end} that perform HOI set prediction in a one-stage manner, the proposed STIP decomposes this process into two phases: the proposed scheme first produces interaction proposals as high-quality interaction queries and then takes them as non-parametric queries to trigger the HOI set prediction. Moreover, this STIP  scheme goes beyond the conventional relational reasoning via vanilla Transformer by leveraging a structure-aware Transformer to exploit the rich inter- or intra-interaction structure, thereby boosting the performance of the HOI detection.

\section{Approach}
In this work, we devise the Structure-aware Transformer over Interaction Proposals (STIP) that casts HOI detection as a set prediction problem in a two-phase fashion. Meanwhile, this scheme boosts HOI set prediction with the prior knowledge of inter- and intra-interaction structures. Figure \ref{fig:framework} depicts an overview of the proposed STIP. The whole framework consists of four main components, i.e., DETR for object detection, interaction proposal network for producing interaction proposals, interaction-centric graph construction, and structure-aware Transformer for HOI set prediction.
Specifically, an off-the-shelf DETR \cite{carion2020end} is first adopted to detect humans and objects within the input image. Next, based on the detection results, we design the Interaction Proposal Network (IPN) to select the most interactive human-object pairs as interaction proposals. After that, we take all selected interaction proposals as graph nodes to construct an interaction-centric graph to reveal the inter-interaction semantic structure and intra-interaction spatial structure. The selected interaction proposals are further taken as non-parametric queries to trigger the HOI set prediction via a structure-aware Transformer through exploiting the structured prior knowledge derived from interaction-centric graph to strengthen relational reasoning.

\subsection{Interaction Proposal Network}
Conditioned on the detected human and object instances from DETR, the Interaction Proposal Network (IPN) targets for producing interaction proposals, i.e., the probably interactive human-object pairs. Concretely, we first construct all possible human-object pairs with pairwise connectivity between detected humans and objects. For each human-object pair, the IPN further predicts the probability of interaction existing in between (i.e., ``interactiveness" score) through a multi-layer perceptron (MLP). Only the top-$K$ human-object pairs with highest interactiveness scores are finally emitted as the output interaction proposals.

\textbf{Human-Object Pairs Construction.}
Here we connect each pair of detected human and object instances, yielding all possible human-object pairs within the input image. Each human-object pair can be represented from three perspectives, i.e., the appearance feature, spatial feature, and linguistic feature of human and object. In particular, the appearance feature is directly represented as the concatenation of human and object instance features derived from DETR (i.e., the 256-dimensional region feature before final prediction heads). By defining the normalized center coordinates of human and object bounding boxes as $(c_x^h, c_y^h)$ and $(c_x^o, c_y^o)$, we measure the spatial feature as the concatenation of all geometric properties, i.e., $[dx, dy, dis, arctan(\frac{dy}{dx}), A_h, A_o, I, U]$, where $dx=c_x^h-c_x^o, dy=c_y^h-c_y^o, dis=\sqrt{dx^2+dy^2}$. $A_h, A_o, I, U$ denote the areas of human, object, their intersection, and union boxes, respectively. The linguistic feature is achieved by encoding the label name of object (one-hot vector) into 300-dimensional vector. The final representation of each human-object pair is calculated as the concatenation of appearance, spatial, and linguistic features.

\begin{figure}[t]
\vspace{-0.32in}
\begin{center}
   \includegraphics[width=0.8\linewidth]{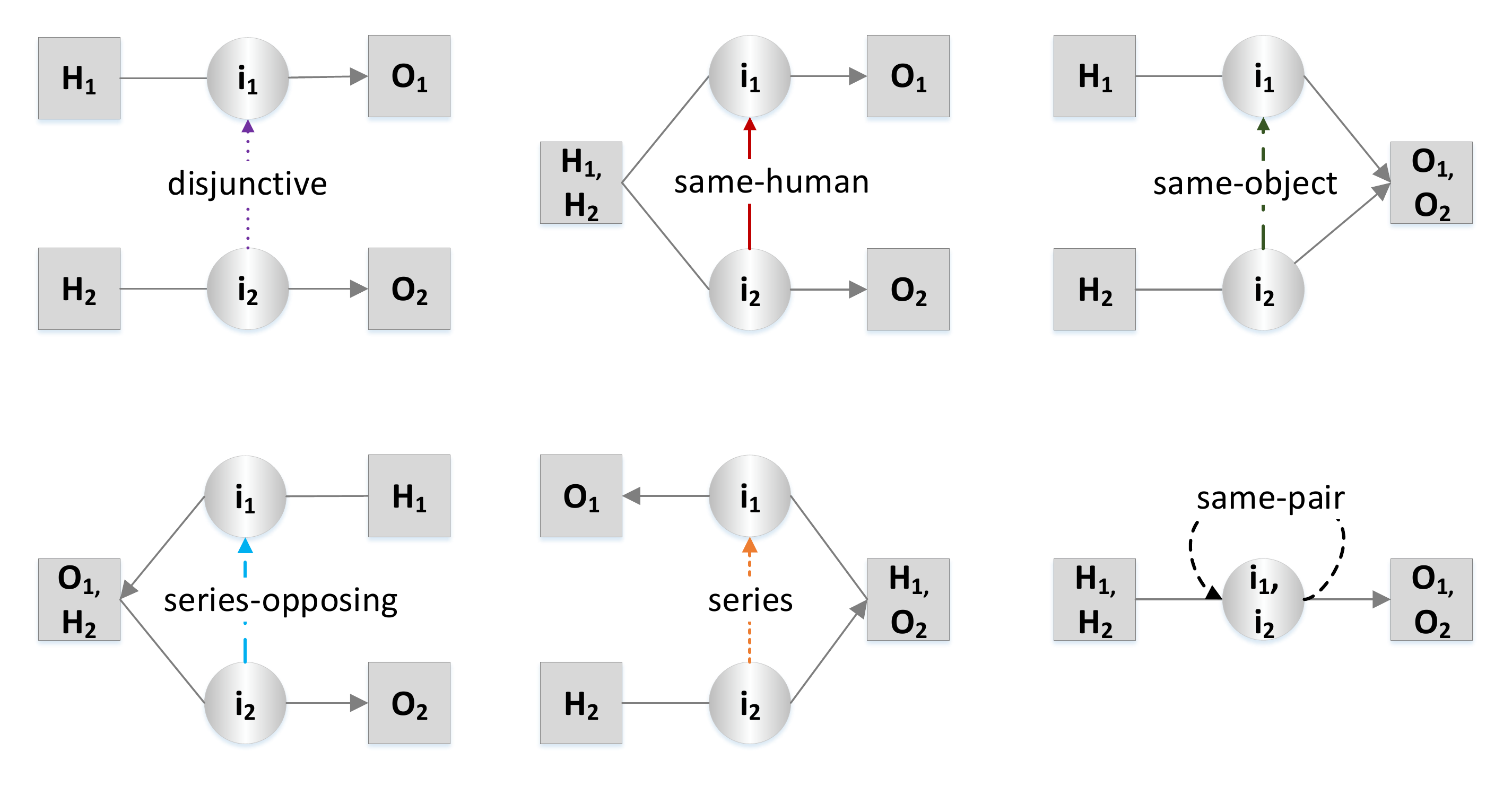}
\end{center}
\vspace{-0.25in}
   \caption{Definition of six kinds of inter-interaction semantic dependencies $\langle$HOI($i_2$)$\rightarrow$HOI($i_1$)$\rangle$ between interaction HOI($i_1$) and HOI($i_2$) (square: human/object instance, circle: interaction).}
\label{fig:self-attention}
\vspace{-0.22in}
\end{figure}

\textbf{Interactiveness Prediction.}
The interactiveness prediction module in IPN takes the feature of each human-object pair as input, and learns to predict the probability whether interactions exist between this pair, i.e., interactiveness score. We frame this sub-task of interactiveness prediction as binary classification problem, and implement this module as MLP coupled with Sigmoid activation. During training, for each input image, we sample at most $K$ human-object pairs, which consist of positive and negative pairs. Note that if both IoUs of predicted human and object bounding boxes in one human-object pair w.r.t ground-truths are larger than 0.5, we treat this pair as positive sample, otherwise it is a negative sample.
One natural way to fetch negative pairs is to use randomly sampling strategy. Instead, here we employ hard mining strategy \cite{shrivastava2016training} to sample negative pairs with high predicted interactiveness scores, aiming to facilitate the learning of interactiveness prediction. After feeding all the $N$ sampled human-object pairs in a mini-batch into interactiveness prediction module, we optimize this module with focal loss \cite{lin2017focal} ($FL$):

\vspace{-0.1in}
\begin{equation}\small
\vspace{-0.0in}
L_{proposal} = \frac{1}{\sum_{i=1}^N z_i} \sum_{i=1}^N FL(\hat{z}_i, z_i),
\label{eq:loss_proposal}
\end{equation}
where $z_i \in \{0,1\}$ indicates whether interactions exist in ground-truth and $\hat{z}_i$ is the predicted interactiveness score.
At inference, only the top-$K$ human-object pairs with highest interactiveness scores are taken as interaction proposals.

\subsection{Interaction-Centric Graph}
Based on all the selected interaction proposals of each input image via IPN, we next present how to construct an interaction-centric graph that fully unfolds the rich prior knowledge of inter- and intra-interaction structures. Technically, we take each interaction proposal as one graph node, and the interaction-centric complete graph is thus built by densely connecting every two nodes as graph edges.

\textbf{Inter-interaction Semantic Structure.}
Intuitively, there exists a natural semantic structure among interactions within a same image. For example, when we find the interaction of ``human hold mouse" in an image, it is very likely that the mentioned ``human'' is associated with another interaction of ``human look-at screen." This motivates us to exploit such common sense knowledge implied in the inter-interaction semantic structure to boost relational reasoning among interactions for HOI detection. Formally, we express the directional semantic connectivity as $\langle$HOI($i_2$)$\rightarrow$HOI($i_1$)$\rangle$, which denotes the relative semantic dependency of interaction proposal HOI($i_1$) against interaction proposal HOI($i_2$). Six kinds of inter-interaction semantic dependencies are thus defined according to whether two interaction proposals share the same human or object instance, as shown in Figure \ref{fig:self-attention}.

Concretely, if HOI($i_1$) and HOI($i_2$) do not share any human/object instance, we classify their dependency as ``$disjunctive$" ({class 0}). If HOI($i_1$) and HOI($i_2$) only share the same human/object instance, we set the label of dependency as ``$same$-$human$" ({class 1}) or ``$same$-$object$" ({class 2}). When the human/object instance of HOI($i_1$) is exactly the object/human instance of HOI($i_2$), the dependency is classified as ``$series$-$opposing$" ({class 3}) and ``$series$" ({class 4}), respectively. If both of the human and object instances of HOI($i_1$) and HOI($i_2$) are same, the label of this dependency is ``$same$-$pair$" ({class 5}).

\begin{figure}[t]
\vspace{-0.35in}
\begin{center}
   \includegraphics[width=0.93\linewidth]{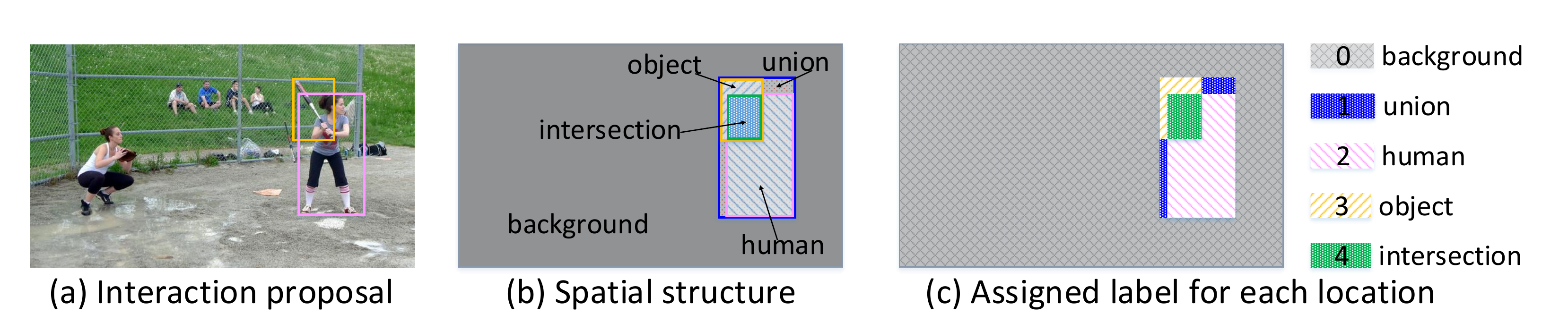}
\end{center}
\vspace{-0.26in}
   \caption{Definition of intra-interaction spatial structure for each interaction: (a) interaction proposal in an image; (b) the spatial structure, i.e., the layout of each component in this interaction; (c) the assigned label for each location in image.}
\label{fig:layout_structure}
\vspace{-0.2in}
\end{figure}

\textbf{Intra-interaction Spatial Structure.}
The inter-interaction semantic structure over the whole interaction-centric graph only unfolds the holistically semantic dependencies across all interaction proposals, while leaving the locally spatial structure of human/object within each interaction proposal unexploited. Therefore, we characterize each graph node with an intra-interaction spatial structure, which can be interpreted as the layout of each component in the corresponding interaction proposal (see Figure \ref{fig:layout_structure}). Specifically, we first identify the spatial location of each component (i.e., $background$, $union$, $human$, $object$, and $intersection$) for this interaction over the whole image, and then assign layout label $l_{ij} \in \{0,1,2,3,4\}$ to each location in this image according to the corresponding component.

\subsection{Structure-aware Transformer}
With the $K$ interaction proposals and the interaction-centric graph, we next present how to integrate the prior knowledge of inter- and intra-interaction structures into relational reasoning for HOI set prediction in STIP. In particular, a structure-aware Transformer is devised to contextually encode all interaction proposals with additional guidance of inter- and intra-interaction structures via structure-aware self-attention and cross-attention modules, yielding structure-aware HOI features for predicting HOI triplets.

\textbf{Preliminary.}
We first briefly recall the widely adopted vanilla Transformer in vision tasks \cite{li2021contextual,li2021scheduled,pan2020auto,pan2020x} that capitalizes on attention mechanism, which aims to transform a sequence of queries $\boldsymbol{q} = (\boldsymbol{q}_1, ..., \boldsymbol{q}_m)$ plus a set of key-value pairs ($\boldsymbol{k}=(\boldsymbol{k}_1,...,\boldsymbol{k}_n), \boldsymbol{v}=(\boldsymbol{v}_1,...,\boldsymbol{v}_n)$) into the output sequence $\boldsymbol{o} = (\boldsymbol{o}_1, ..., \boldsymbol{o}_m)$. Each output element $\boldsymbol{o}_i$ is computed by aggregating all values weighted with attention: $\boldsymbol{o}_i = \sum_{j} \alpha_{ij} (\boldsymbol{W}_v \boldsymbol{v}_j)$, where each attention weight $\alpha_{ij}$ is normalized with softmax ($\alpha_{ij} = \frac{exp(e_{ij})}{\sum_{j} exp(e_{ij})}$). Here the primary attention weight $e_{ij}$ is measured as the scaled dot-product between each key $\boldsymbol{k}_j$ and query $\boldsymbol{q}_i$:
\vspace{-0.05in}
\begin{equation}\small
e_{ij} = \frac{(\boldsymbol{W}_q \boldsymbol{q}_i)^T (\boldsymbol{W}_k \boldsymbol{k}_j)}{\sqrt{d_{key}}}.
\label{eq:unnorm_attention-weight}
\end{equation}
Note that $d_{key}$ is the dimension of keys, and $\boldsymbol{W}_q, \boldsymbol{W}_k, \boldsymbol{W}_v$ are learnable embedding matrices.

\textbf{Structure-aware Self-attention.}
Existing Transformer-type HOI detectors perform relational reasoning among interactions via self-attention module in vanilla Transformer for HOI set prediction. However, the relational reasoning process in vanilla Transformer is triggered by the parametric interaction queries, and leaves the prior knowledge of inter-interaction structure under-exploited. As an alternative, our structure-aware Transformer starts HOI set prediction from the non-parametric queries (i.e., the selected interaction proposals), and further upgrades the conventional relation reasoning with inter-interaction semantic structure through structure-aware self-attention module.

Specifically, by taking the $K$ interaction proposals $\boldsymbol{q}$ as interaction queries, keys, and values, the structure-aware self-attention module conducts the inter-interaction structure-aware reasoning among interactions to strengthen the HOI representation of each interaction.
Inspired by relative position encoding in \cite{shaw2018self}, we supplement each key $\boldsymbol{q}_j$ with the encodings of its inter-interaction semantic dependency with regard to query $\boldsymbol{q}_i$, which is measured as the concatenation of $\boldsymbol{q}_j$ and the corresponding semantic dependency label $d_{ij} \in \{0,1...,5\}$. In this way, we incorporate the inter-interaction semantic structure into the learning of attention weight by modifying Eq. (\ref{eq:unnorm_attention-weight}) as:
\begin{equation}\small
e_{ij}^{self} = \frac{(\boldsymbol{W}_q \boldsymbol{q}_i)^T (\boldsymbol{W}_k \boldsymbol{q}_j + \boldsymbol{\psi}(\boldsymbol{q}_j, \boldsymbol{E}_{dep}(d_{ij})))}{\sqrt{d_{key}}},
\end{equation}
where $\boldsymbol{E}_{dep}$ denotes the embedding matrix of semantic dependency label and $\boldsymbol{\psi}$ is implemented as a 2-layer MLP to encode the inter-interaction semantic dependency.
Accordingly, the output intermediate HOI features $\hat{\boldsymbol{q}}$ of structure-aware self-attention module are endowed with the holistically semantic structure among interactions.

\begin{figure}[t]
\vspace{-0.35in}
\begin{center}
   \includegraphics[width=0.92\linewidth]{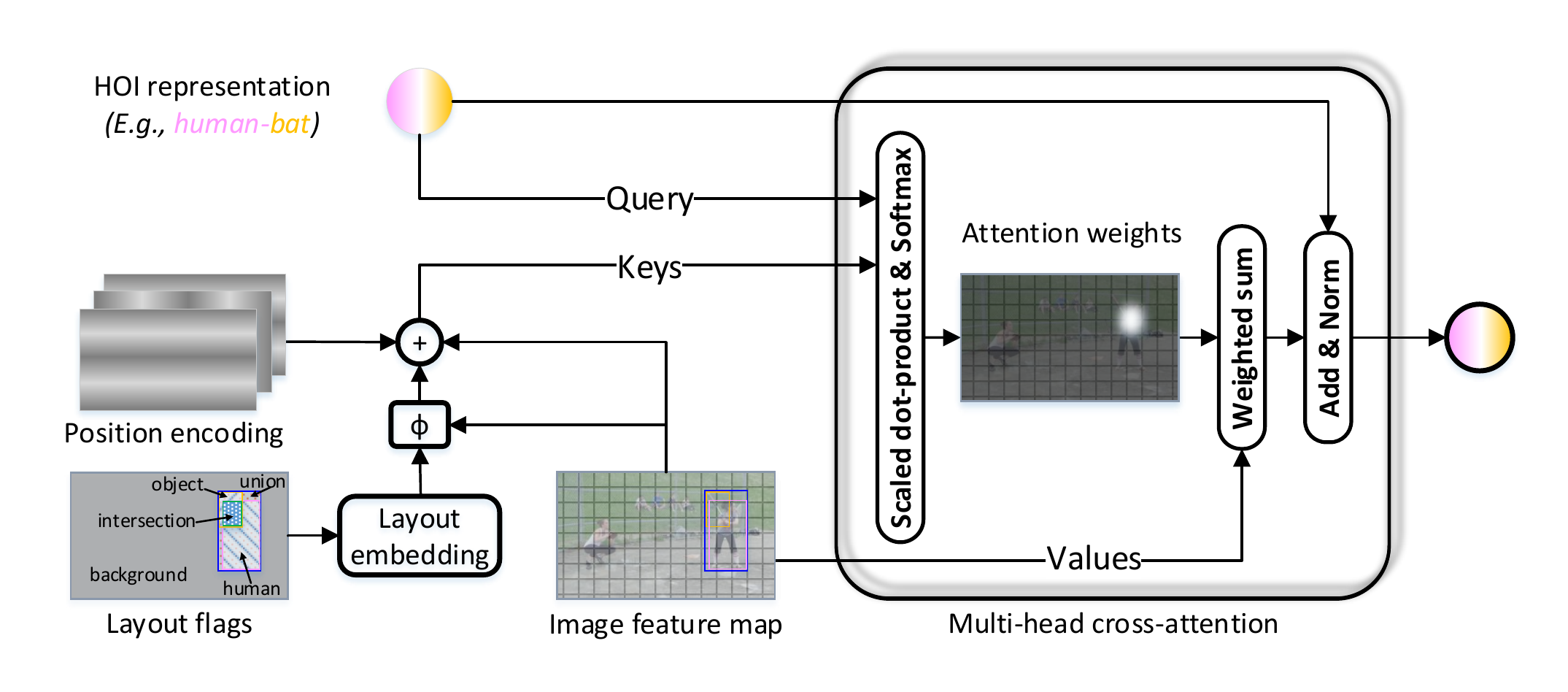}
\end{center}
\vspace{-0.3in}
   \caption{Structure-aware cross-attention module.}
   \vspace{-0.26in}
\label{fig:cross-attention}
\end{figure}

\textbf{Structure-aware Cross-attention.}
Next, based on the intermediate HOI features $\hat{\boldsymbol{q}}$, a structure-aware cross-attention module (see Figure \ref{fig:cross-attention}) is utilized to further enhance HOI features by exploiting contextual information between interactions and the original image feature map in DETR. Formally, we take the $K$ intermediate HOI features $\hat{\boldsymbol{q}}=(\boldsymbol{\hat{q}}_1, ..., \boldsymbol{\hat{q}}_K)$ as queries, and the image feature map $\boldsymbol{x} = (\boldsymbol{x}_1, ..., \boldsymbol{x}_n)$ as keys and values. The structure-aware cross-attention module performs the intra-interaction structure-aware reasoning over the image feature map to strengthen the HOI feature of each interaction. Similar to structure-aware self-attention module, each key $\boldsymbol{x}_j$ is supplemented with the encodings of the intra-interaction spatial structure with regard to query $\boldsymbol{\hat{q}}_i$ (i.e., the concatenation of $\boldsymbol{x}_j$ and its assigned layout label $l_{ij} \in \{0,1,2,3,4\}$). The learning of attention weight in structure-aware cross-attention module is thus integrated with the intra-interaction spatial structure, which is measured as:
\begin{equation}\small
e_{ij}^{cross} = \frac{(\boldsymbol{W}_{\hat{q}} {\boldsymbol{\hat{q}}_i})^T (\boldsymbol{W}_{\hat{k}} \boldsymbol{x}_j + \boldsymbol{pos}_j +\boldsymbol{\phi}(\boldsymbol{x}_j, \boldsymbol{E}_{lay}(l_{ij})))}{\sqrt{d_{key}}},
\end{equation}
where $\boldsymbol{pos}_j$ is the position encoding, $\boldsymbol{E}_{lay}$ is the embedding matrix of layout label, and we implement $\boldsymbol{\phi}$ as a 2-layer MLP to encode the intra-interaction spatial structure.

\subsection{Training Objective}
During training, we feed the final output HOI representations of structure-aware Transformer into the interaction classifier (implemented as a 2-layer MLP) to predict the interaction classes of each interaction proposal. The objective of interaction classification is measured via focal loss:
\begin{equation}
L_{cls} = \frac{1}{\sum_{i=1}^N  \sum_{c=1}^C y_{ic}} \sum_{i=1}^N \sum_{c=1}^C FL(\hat{y}_{ic}, y_{ic}),
\label{eq:loss_classification}
\end{equation}
where $C$ is the number of interaction classes, $y_{ic} \in \{0,1\}$ indicates whether the labels of $i$-th proposal contain the $c$-th interaction class, and $\hat{y}_{ic}$ is the predicted probability of $c$-th interaction class. Accordingly, the overall objective of our STIP integrates the interactiveness prediction objective in Eq. (\ref{eq:loss_proposal}) and interaction classification objective in Eq. (\ref{eq:loss_classification}):
\begin{equation}
L_{STIP} = L_{proposal} + L_{cls}.
\end{equation}

\section{Experiments}
Here we empirically evaluate STIP on two common HOI detection datasets, i.e., V-COCO \cite{gupta2015visual} and HICO-DET \cite{chao2018learning}.

\subsection{Datasets and Experimental Settings}

\textbf{V-COCO} is a popular dataset for benchmarking HOI detection, which is a subset of MS-COCO \cite{lin2014microsoft} covering 29 action categories. This dataset consists of 2,533 training images, 2,867 validation images, and 4,946 testing images. Following the settings in \cite{kim2021hotr}, we adopt Average Precision ($AP_{role}$) over 25 interactions as evaluation metric. Two kinds of $AP_{role}$, i.e., \textbf{$AP_{role}^{\#1}$} and \textbf{$AP_{role}^{\#2}$}, are reported under two scenarios with different scoring criterions for object occlusion cases. Specifically, in the scenario of $AP_{role}^{\#1}$, the model should manage to infer the occluded object correctly by predicting the 2D location of its bounding box as [0,0,0,0], meanwhile precisely localizing the corresponding human bounding box and recognizing the interaction in between. In contrast, for the scenario of $AP_{role}^{\#2}$, there is no need to infer the occluded object.

\textbf{HICO-DET} is a larger HOI detection benchmark, which contains 37,536 and 9,515 images for training and testing, respectively. The whole dataset covers 600 categories of {\emph{$\langle human, object, interaction\rangle$}} triplets, covering the same 80 object categories as in MS-COCO \cite{lin2014microsoft} and 117 verb categories. We follow \cite{chao2018learning} and report mAP in two different settings (\textbf{$Default$} and \textbf{$Known\: Object$}). Here the $Default$ setting represents that the mAP is calculated over all testing images, while $Known\: Object$ measures the AP of each object solely over the images containing that object class.
For each setting, we report the AP over three different HOI category sets, i.e., \textbf{Full} (all 600 HOI categories), \textbf{Rare} (138 HOI categories where each one contains less than 10 training samples), and \textbf{Non-Rare} (462 HOI categories where each one contains 10 or more training samples).

\textbf{Implementation Details.}
For fair comparison with state-of-the-art baselines, we adopt the same object detector DETR pre-trained over MS-COCO (backbone: ResNet-50) and all learnable parameters in DETR are frozen during training as in \cite{kim2021hotr}. On HICO-DET dataset, we additionally report the results by fine-tuning DETR on HICO-DET and the performances by further jointly fine-tuning object detector and HOI detector. In the experiments, we select the top-32 probably interactive human-object pairs as the output interaction proposals of Interaction Proposal Network. Our proposed structure-aware Transformer consists of 6 stacked layers (structure-aware self-attention plus cross-attention modules). The whole architecture is trained over 2 Nvidia 2080ti GPUs with AdamW optimizer. The mini-batch size is 8 and we set the initial learning rate as $5\times10^{-5}$. The maximum training epoch number is 30.

\begin{table}[t]
\begin{center}
\vspace{-0.2in}
\scalebox{0.85}{
\begin{tabular}{lllcc}
\hline
Method & Backbone & Feature & $AP^{\#1}_{role}$ & $AP^{\#2}_{role}$ \\
\hline\hline
\emph{One-stage methods} &&& \\
UnionDet \cite{kim2020uniondet} & R50 &A& 47.5 & 56.2 \\
IPNet \cite{wang2020learning} & HG-104 &A& 51.0 & - \\
GGNet \cite{zhong2021glance} & HG-104 &A& 54.7 & - \\
HOITrans \cite{zou2021end} & R50 &A& 52.9 & - \\
AS-Net \cite{chen2021reformulating} & R50 &A&  53.9 & - \\
HOTR \cite{kim2021hotr} & R50 &A& 55.2 & 64.4 \\
QPIC \cite{tamura2021qpic} & R50 &A& 58.8 & 61.0 \\
\hline
\emph{Two-stage methods} &&& \\
InteractNet \cite{gkioxari2018detecting} & R50-FPN &A& 40.0 & 48.0 \\
GPNN \cite{qi2018learning}   & R101 &A& 44.0 & - \\
TIN \cite{li2019transferable} & R50 &A+S+P& 48.7 & - \\
DRG \cite{gao2020drg}   & R50-FPN &A+S+L& 51.0 & - \\
FCMNet \cite{liu2020amplifying}  & R50 &A+S+L+P& 53.1 & - \\
ConsNet \cite{liu2020consnet}   & R50-FPN &A+S+L& 53.2 & - \\
IDN \cite{li2020hoi} & R50 &A+S& 53.3 & 60.3 \\
\rowcolor{Gray}
STIP (Ours) & R50 &A& \textbf{65.1} & \textbf{69.7} \\
\rowcolor{Gray}
STIP (Ours) & R50 &A+S+L& \textbf{66.0} & \textbf{70.7} \\
\hline
\end{tabular}}
\end{center}
\vspace{-0.2in}
\caption{Performance comparison on V-COCO dataset. The letters in Feature column indicate the input features: \textbf{A} (Appearance/Visual features), \textbf{S} (Spatial features \cite{gao2018ican}), \textbf{L} (Linguistic feature of label semantic embeddings), \textbf{P} (Human pose feature).}
\label{tab:vcoco-res}
\vspace{-0.2in}
\end{table}

\begin{table*}[!htb]\small
\begin{center}
\vspace{-0.22in}
\scalebox{0.88}{
\begin{tabular}{lllccccccc}
\hline
\multirow{3}{*}{Method} & \multirow{3}{*}{Backbone} & \multirow{3}{*}{Feature} & \multicolumn{3}{c}{Default} && \multicolumn{3}{c}{Known Object} \\
\cmidrule{4-6} \cmidrule{8-10}
&&& Full & Rare & Non-Rare && Full & Rare & Non-Rare \\
\hline\hline
\multicolumn{5}{l}{\emph{Object detector pre-trained on MS-COCO}} \\
InteractNet \cite{gkioxari2018detecting} & R50-FPN &A& 9.94 & 7.16 & 10.77 && - & - & - \\
GPNN \cite{qi2018learning}   & R101 &A& 13.11 & 9.41 & 14.23 && - & - & -\\
UnionDet \cite{kim2020uniondet} & R50 &A& 14.25 & 10.23 & 15.46 && 19.76 & 14.68 & 21.27\\
TIN \cite{li2019transferable}  & R50 &A+S+P& 17.22 & 13.51 & 18.32 && 19.38 & 15.38 & 20.57 \\
IPNet \cite{wang2020learning} & R50-FPN &A& 19.56 & 12.79 & 21.58 && 22.05 & 15.77 & 23.92 \\
DRG \cite{gao2020drg}   & R50-FPN &A+S+L& 19.26 & 17.74 & 19.71 && 23.40 & 21.75 & 23.89 \\
FCMNet \cite{liu2020amplifying} & R50 &A+S+L+P& 20.41 & 17.34 & 21.56 && 22.04 & 18.97 & 23.12 \\
ConsNet \cite{liu2020consnet}  & R50-FPN &A+S+L& 22.15 & 17.12 & 23.65 && - & - & - \\
IDN \cite{li2020hoi} & R50 &A+S& 23.36 & 22.47 & 23.63 && 26.43 & 25.01 & 26.85 \\
HOTR \cite{kim2021hotr}  & R50 &A& 23.46 & 16.21 & 25.60 && - & - & - \\
AS-Net \cite{chen2021reformulating}  & R50 &A& 24.40 & 22.39 & 25.01 && 27.41 & 25.44 & 28.00 \\
\rowcolor{Gray}
STIP (Ours) & R50 &A& 28.11 & 25.85 & 28.78 && 31.23 & 27.93 & 32.22 \\
\rowcolor{Gray}
STIP (Ours) & R50 &A+S+L& \textbf{28.81} & \textbf{27.55} & \textbf{29.18} && \textbf{32.28} & \textbf{31.07} & \textbf{32.64}\\
\hline
\multicolumn{5}{l}{\emph{Object detector fine-tuned on HICO-DET}} \\
DRG \cite{gao2020drg}    & R50-FPN &A+S+L& 24.53 & 19.47 & 26.04 && 27.98 & 23.11 & 29.43\\
ConsNet \cite{liu2020consnet}  & R50-FPN &A+S+L& 24.39 & 17.10 & 26.56 && - & - & -\\
IDN \cite{li2020hoi}  & R50 &A+S& 26.29 & 22.61 & 27.39 && 28.24 & 24.47 & 29.37 \\
HOTR   \cite{kim2021hotr} & R50 &A& 25.10 & 17.34 & 27.42 && - & - & -\\
\rowcolor{Gray}
STIP (Ours) & R50 &A& 29.76 & 26.94 & 30.61 && 32.84 & 29.05 & 33.85\\
\rowcolor{Gray}
STIP (Ours) & R50 &A+S+L& \textbf{30.56} & \textbf{28.15} & \textbf{31.28} && \textbf{33.54} & \textbf{30.93} & \textbf{34.31}\\
\hline
\multicolumn{5}{l}{\emph{Jointly fine-tune object detector \& HOI detector on HICO-DET}} \\
UnionDet \cite{kim2020uniondet}  & R50 &A& 17.58 & 11.72 & 19.33 && 19.76 & 14.68 & 21.27 \\
PPDM \cite{liao2020ppdm}  & HG104  &A& 21.73 & 13.78 & 24.10 && 24.58 & 16.65 & 26.84\\
GGNet \cite{zhong2021glance} & HG104 &A& 29.17 & 22.13 & 30.84 && 33.50 & 26.67 & 34.89\\
HOITrans \cite{zou2021end} & R50 &A& 23.46 & 16.91 & 25.41 && 26.15 & 19.24 & 28.22 \\
AS-Net \cite{chen2021reformulating}  & R50 &A& 28.87 & 24.25 & 30.25 && 31.74 & 27.07 & 33.14\\
QPIC \cite{tamura2021qpic} & R50 &A& 29.07 & 21.85 & 31.23 && 31.68 & 24.14 & 33.93\\
\rowcolor{Gray}
STIP (Ours) & R50 &A& 31.60 & 27.75 & 32.75 && 34.41 & 30.12 & 35.69 \\
\rowcolor{Gray}
STIP (Ours) & R50  &A+S+L& \textbf{32.22} & \textbf{28.15} & \textbf{33.43} && \textbf{35.29} & \textbf{31.43} & \textbf{36.45} \\
\hline
\end{tabular}}
\end{center}
\vspace{-0.2in}
\caption{Performance comparison on HICO-DET dataset. The letters in Feature column indicate the input features: \textbf{A} (Appearance/Visual features), \textbf{S} (Spatial features \cite{gao2018ican}), \textbf{L} (Linguistic feature of label semantic embeddings), \textbf{P} (Human pose feature).}
\label{tab:hicodet-res}
\vspace{-0.26in}
\end{table*}

\subsection{Performance Comparisons}
\textbf{V-COCO.} Table \ref{tab:vcoco-res} summarizes the performance comparisons in terms of $AP_{role}^{\#1}$ and $AP_{role}^{\#2}$ on V-COCO. In general, the results across all metrics under the same backbone (ResNet-50, R50 in short) consistently demonstrate that our STIP exhibits better performances against existing techniques, including both one-stage methods (e.g., UnionDet, AS-Net, HOTR, and QPIC) and two-stage methods (e.g., FCMNet, ConsNet, and IDN). The results generally highlight the key advantage of two-phase HOI set prediction and the exploitation of inter- and intra-interaction structures. In particular, the conventional two-stage HOI detectors (e.g., GPNN, TIN, DRG) commonly construct instance-centric graph to mine contextual information among instances. Instead, recent Transformer-style HOI detectors (e.g., HOITrans, AS-Net, HOTR, QPIC) fully capitalize on vanilla Transformer to perform relational reasoning among instances/interactions, thereby leading to performance boosts. However, when only using appearance features (A), the $AP_{role}^{\#1}$ and $AP_{role}^{\#2}$ of HOTR and QPIC are still lower than our STIP, which not only takes interaction proposals as non-parametric interaction queries to trigger HOI set prediction, but also leverages a structure-aware Transformer to exploit the prior knowledge of inter-interaction and intra-interaction structures. For our STIP, a further performance improvement is attained when utilizing more kinds of features (e.g., spatial and linguistic features).

\begin{table}[!tb]
\begin{center}
\scalebox{0.72}{
\begin{tabular}{lcccccc}
\hline
\multirow{3}{*}{Method} & \multicolumn{2}{c}{V-COCO} && \multicolumn{3}{c}{HICO-DET (Default)} \\
\cmidrule{2-3} \cmidrule{5-7}
   & $AP^{\#1}_{role}$ & $AP^{\#2}_{role}$ && Full & Rare & Non-Rare \\
\hline \hline
Base & 52.49 & 58.25 && 21.74 & 18.09 & 22.83 \\
\hline
\:\:+HM & 58.45 & 62.64 && 24.16 & 19.45 & 25.57 \\
\:\:+HM+TR & 63.50 & 68.07 && 28.62 & 26.09 & 29.38 \\
\:\:+HM+TR\textsuperscript{\emph{SS}} & 64.99 & 69.94 && 29.65 & 26.52 & 30.59 \\
\:\:+HM+TR\textsuperscript{\emph{SC}} & 65.04 & 69.76 && 29.74 & 27.07 & 30.54 \\
\:\:+HM+TR\textsuperscript{\emph{SS+SC}} (\textbf{STIP}) & \textbf{66.04} & \textbf{70.65} && \textbf{30.56} & \textbf{28.15} & \textbf{31.28}  \\
\hline
\end{tabular}}
\end{center}
\vspace{-0.22in}
\caption{Performance contribution of each component in our STIP. \textbf{HM}: Hard Mining strategy for training interaction proposal network. \textbf{TR}: vanilla TRansformer. \textbf{TR\textsuperscript{\emph{SS}}}: TRansformer with only Structure-aware Self-attention that exploits inter-interaction structure. \textbf{TR\textsuperscript{\emph{SC}}}: TRansformer with only Structure-aware Cross-attention that exploits intra-interaction structure.}
\label{tab:main-ablation}
\vspace{-0.25in}
\end{table}

\textbf{HICO-DET.} We further evaluate our STIP on the more challenging HICO-DET dataset. Table \ref{tab:hicodet-res} reports the mAP scores over three different HOI category sets for each setting (Default/Known Object) in comparison with the state-of-the-art methods. Here we include three different training settings, i.e., pre-train object detector on MS-COCO, fine-tune object detector on HICO-DET, and jointly fine-tune object detector and HOI detector on HICO-DET, for fair comparison. Similar to the observations on V-COCO, our STIP achieves consistent performance gains against existing HOI detectors across all the metrics for each training setting. The results basically demonstrate the advantage of triggering HOI set prediction with the non-parametric interaction proposals and meanwhile exploiting the holistically semantic structure among interaction proposals \& the locally spatial structure within each interaction proposal.

\subsection{Experimental Analysis}
\textbf{Ablation Study.}
To examine the impact of each design in STIP, we conduct ablation study by comparing different variants of STIP on V-COCO and HICO-DET datasets in Table \ref{tab:main-ablation}. Note that all experiments on HICO-DET here are conducted under the training setting of object detector fine-tuned on HICO-DET. We start from the basic model (\textbf{Base}), which utilizes a basic interaction proposal network (randomly sampling negative samples for training, without hard mining strategy). The generated interaction proposals in Base model are directly leveraged for interaction classification, without any Transformer-style structure for boosting HOI prediction. Next, we extend Base model by leveraging hard mining strategy to select the hard negative human-object pairs with higher interactiveness scores for training interaction proposal network, yielding \textbf{Base+HM} which achieves better performances. After that, by additionally involving a vanilla Transformer to perform relational reasoning among interaction proposals, another variant of our model (\textbf{Base+HM+TR}) leads to performance improvements across all metrics. Furthermore, we upgrade the vanilla Transformer with structure-aware self-attention that exploits the holistically semantic structure among interaction proposals, and this ablated run (\textbf{Base+HM+TR\textsuperscript{\emph{SS}}}) outperforms Base+HM+TR. Meanwhile, the vanilla Transformer can be upgraded with structure-aware cross-attention that exploits the locally spatial structure within each interaction proposal, and \textbf{Base+HM+TR\textsuperscript{\emph{SC}}} also exhibits better performances. These observations basically validate the merit of exploiting the structured prior knowledge, i.e., inter-interaction or intra-interaction structure, for HOI detection. Finally, when jointly upgrading the vanilla Transformer with structure-aware self-attention and structure-aware cross-attention (i.e., our \textbf{STIP}), the highest performances are attained.

\begin{table}[!tb]
\begin{center}
\vspace{-0.22in}
\scalebox{0.73}{
\begin{tabular}{p{3.4cm}cccccc}
\hline
\multirow{3}{3.4cm}{\centering \small \# of selected interaction proposals ($K$)} & \multicolumn{2}{c}{V-COCO} && \multicolumn{3}{c}{HICO-DET (Default)} \\
\cmidrule{2-3} \cmidrule{5-7}
   & $AP^{\#1}_{role}$ & $AP^{\#2}_{role}$ && Full & Rare & Non-Rare \\
\hline\hline
\centering{8} & 64.20 & 69.11 && 29.03 & 28.16 & 29.29 \\
\centering{16} & 65.68 & 70.63 && 30.18 & 28.66 & 30.64 \\
\centering{\underline{32}} & \textbf{66.04} & \textbf{70.65} && 30.56 & 28.15 & \textbf{31.28} \\
\centering{64} & 65.93 & 70.50 && \textbf{30.72} & \textbf{28.96} & 31.24 \\
\centering{100} & 65.78 & 70.45 && 30.40 & 27.89 & 31.14 \\
\hline
\end{tabular}}
\end{center}
\vspace{-0.2in}
\caption{Performance comparison by using different number of selected interaction proposals ($K$) for interaction-centric graph construction in our STIP.}
\label{tab:nqueries-ablation}
\vspace{-0.25in}
\end{table}

\textbf{Effect of Selected Interaction Proposal Number $K$ for Interaction-centric Graph Construction.}
Recall that the interaction proposal network in our STIP selects only the top-$K$ human-object pairs with highest interactiveness scores as the output interaction proposals for constructing the interaction-centric graph. Such $K$ selected interaction proposals are also taken as non-parametric interaction queries to trigger HOI set prediction in the structure-aware Transformer. Here we vary $K$ from 8 to 100 to explore the relationship between the performance and the select interaction proposal number $K$. As shown in Table \ref{tab:nqueries-ablation}, the best performances across most metrics are attained when $K$ is set as 32. In particular, enlarging the number of selected interaction proposals (until $K=32$) can generally lead to performance boosts on two datasets. Once $K$ is larger than 64, the performances slightly decrease. We speculate that the increase of selected interaction proposals result in more invalid proposals, which may affect the overall stability of relational reasoning among interaction proposals. Accordingly, we empirically set $K$ as 32.

\begin{table}[!tb]
\begin{center}
\vspace{-0.22in}
\scalebox{0.8}{
\begin{tabular}{p{2.2cm}cccccc}
\hline
\multirow{3}{2.2cm}{\centering \small \# of layers ($L$)} & \multicolumn{2}{c}{V-COCO} && \multicolumn{3}{c}{HICO-DET (Default)} \\
\cmidrule{2-3} \cmidrule{5-7}
   & $AP^{\#1}_{role}$ & $AP^{\#2}_{role}$ && Full & Rare & Non-Rare \\
\hline\hline
\centering{0} & 58.45 & 62.64 && 24.16 & 19.45 & 25.57 \\
\centering{1} & 64.83 & 69.57 && 29.21 & 26.37 & 30.06 \\
\centering{2} & 65.55 & 70.39 && 30.02 & 28.11 & 30.59 \\
\centering{4} & 66.02 & 70.61 && 30.47 & 29.28 & 30.83 \\
\centering{\underline{6}} & \textbf{66.04} & \textbf{70.65} && 30.56 & 28.15 & \textbf{31.28} \\
\centering{8} & 65.44 & 70.11 && \textbf{30.93} & \textbf{29.78} & 31.27 \\
\hline
\end{tabular}}
\end{center}
\vspace{-0.2in}
\caption{Performance comparison with different layer numbers of the structure-aware Transformer in our STIP.}
\label{tab:Transformer-layer-ablation}
\vspace{-0.25in}
\end{table}

\textbf{Effect of Layer Number $L$ in Structure-aware Transformer.}
To explore the effect of layer number $L$ in structure-aware Transformer, we show the performances on two benchmarks by varying this number from 0 to 8. The best performances across most metrics are achieved when the layer number is set to $L=6$. Specifically, in the extreme case of $L=0$, no self-attention and cross-attention module is utilized, and the model degenerates to a Base+HM model that directly performs interaction classification over interaction proposals without any relational reasoning via Transformer-style structure. When increasing the layer number in structure-aware Transformer, the performances are gradually increased in general. This basically validates the effectiveness of enabling relational reasoning among interaction proposals through structure-aware Transformer. In practice, the layer number $L$ is generally set to~6.

\textbf{Time Analysis.}
We evaluate the inference time of our STIP on a single Nvidia 2080ti GPU by constructing each batch with single testing image. Specifically, for each input batch, object detection via DETR, interaction proposal generation through interaction proposal network, HOI set prediction with structure-aware Transformer, and the other processing (e.g., data loading) takes 41.9ms, 7.8ms, 20.4ms, and 3.8ms, respectively. Consequently, the overall inference stage of STIP finishes in 73.9ms on average, which is comparable to existing one-stage Transformer-style HOI detectors (e.g., the inference time of AS-Net \cite{chen2021reformulating} is 71ms).

\section{Conclusion and Discussion}

In this paper, we have presented STIP, a new end-to-end Transformer-style model for human-object interaction detection. Instead of performing HOI set prediction derived from parametric interaction queries in a one-stage manner, the proposed STIP capitalizes on a two-phase solution for HOI detection by first producing interaction proposals and then taking them as non-parametric interaction queries to trigger HOI set prediction. Furthermore, by going beyond the commonly adopted vanilla Transformer, a novel structure-aware Transformer is designed to exploit two kinds of structured prior knowledge, i.e., inter- and intra-interaction structures, to further boost HOI set prediction. We validate the proposed scheme and analysis through extensive experiments conducted on V-COCO and HICO-DET datasets. More importantly, the proposed STIP achieves new state-of-the-art results on both benchmarks.

\textbf{Broader Impact.} STIP has high potential impact in human-centric applications, such as sports analysis and self-driving vehicles. However, such HOI detection technology can be deployed in human monitoring and surveillance as well which might raise ethical and privacy issues.

{\small
\bibliographystyle{ieee_fullname}
\bibliography{STIP_refs}
}

\end{document}